\pdfoutput=1

\documentclass[11pt]{article}

\usepackage[final]{acl}
\usepackage[]{acl}

\usepackage{times}
\usepackage{latexsym}
\usepackage{longtable}

\usepackage[T1]{fontenc}
\usepackage{multicol}

\usepackage{CJKutf8}

\usepackage{microtype}

\usepackage{hyperref}
\usepackage{amsmath} 
\usepackage{amssymb} 
\usepackage{graphicx} 
 \usepackage{xcolor}
\usepackage{threeparttable}
\usepackage{pifont}
\usepackage{multirow}
\usepackage{subfigure}
\usepackage{graphicx}
\usepackage{booktabs}
\usepackage{enumitem}
\usepackage{algorithm}
\usepackage{algpseudocode}
\usepackage{amsmath}
\usepackage{graphicx} 
\usepackage{mdframed}
\usepackage[utf8]{inputenc}
\usepackage{relsize}
\usepackage{booktabs}
\usepackage{graphicx}
\usepackage{array}
\usepackage{colortbl} 
\usepackage{array}    

 \usepackage{amsmath}

\usepackage{csquotes}
\usepackage{epigraph}
\usepackage{multirow} 
\usepackage{geometry}
\usepackage{array}
\usepackage{longtable}
\usepackage{float}
\usepackage{mdframed}
\usepackage{xcolor}
\definecolor{deepyellow}{rgb}{0.8, 0.7, 0.0} 
\definecolor{deepgreen}{rgb}{0.0, 0.7, 0.0} 

\definecolor{chatgpt_c}{RGB}{121,147,210}
\definecolor{src_c}{RGB}{238,154,189}
\definecolor{tgt_c}{RGB}{139,212,209}
\definecolor{src_tgt_c}{RGB}{149,149,149}
\definecolor{LAT_c}{RGB}{247,182,95}

\definecolor{chinese_red}{RGB}{230,239,255}
\definecolor{chinese_red_small}{RGB}{252,255,230}
\definecolor{chinese_brown}{RGB}{246,230,255}
 \usepackage{booktabs}

\usepackage{graphicx}
\definecolor{win}{RGB}{165,127,183} 
\definecolor{tie}{RGB}{204,161,189}
\definecolor{loss}{RGB}{229,207,221}

\usepackage{CJK}

\usepackage{tcolorbox}
\tcbuselibrary{breakable} 

\title{Evaluating o1-Like LLMs: Unlocking Reasoning for Translation through Comprehensive Analysis}

\author{
Andong Chen\footnotemark[1]\hspace{0.5mm}, 
 Yuchen Song\footnotemark[1]\hspace{0.5mm}, 
 Wenxin Zhu\footnotemark[1]\hspace{0.5mm},
 \textbf{Kehai Chen}\hspace{0.5mm},
 \textbf{Muyun Yang}\hspace{0.5mm},
 \textbf{Tiejun Zhao}\hspace{0.5mm}, 
 \textbf{Min Zhang}\hspace{0.2mm}\hspace{1.5mm} \\
School of Computer Science and Technology, Harbin Institute of Technology, China\\
  ands691119@gmail.com,  \{2021113318, 2021111266\}@stu.hit.edu.cn  \\
  \{chenkehai, yangmuyun, tjzhao, zhangmin2021\}@hit.edu.cn, 
}

\begin{document}
\begin{CJK}{UTF8}{gbsn}
\maketitle
\renewcommand{\thefootnote}{\fnsymbol{footnote}}
\footnotetext[1]{These authors contributed equally to this work and should be considered co-first authors.}
\renewcommand{\thefootnote}{\arabic{footnote}}
\begin{abstract}

The o1-Like LLMs are transforming AI by simulating human cognitive processes, but their performance in multilingual machine translation (MMT) remains underexplored. This study examines: (1) how o1-Like LLMs perform in MMT tasks and (2) what factors influence their translation quality. We evaluate multiple o1-Like LLMs and compare them with traditional models like ChatGPT and GPT-4o. Results show that o1-Like LLMs establish new multilingual translation benchmarks, with DeepSeek-R1 surpassing GPT-4o in contextless tasks. They demonstrate strengths in historical and cultural translation but exhibit a tendency for \textbf{rambling issues} in Chinese-centric outputs. Further analysis reveals three key insights: (1) High inference costs and slower processing speeds make complex translation tasks more resource-intensive. (2) Translation quality improves with model size, enhancing commonsense reasoning and cultural translation. (3) The temperature parameter significantly impacts output quality—lower temperatures yield more stable and accurate translations, while higher temperatures reduce coherence and precision.~\footnote{Our code will be made available at \url{https://github.com/anonymous}.}
\end{abstract}

%

\section{Introduction}

Large Language Models (LLMs) have demonstrated significant potential in the field of Machine Translation (MT)~\cite{DBLP:conf/naacl/ZhuLDXHKCL24,DBLP:conf/emnlp/Peng0ZS0ZOT23,DBLP:journals/corr/abs-2301-08745,DBLP:journals/corr/abs-2302-09210,tyen2023llms,liang2023encouraging,DBLP:journals/corr/abs-2303-16104,DBLP:journals/corr/abs-2308-14186,chen-etal-2024-dual}. Currently, o1-Like LLMs—such as OpenAI’s o1 model~\cite{openai2024reasoning} and its replicated versions like QwQ~\cite{qwen2024qwq}, Marco-o1~\cite{DBLP:journals/corr/abs-2411-14405}, and DeepSeek-R1~\cite{guo2025deepseek} are profoundly transforming the AI landscape, enabling models to handle increasingly complex tasks. These models are renowned for their strong reasoning capabilities, exhibiting human-like deep thinking in extended tests, exploring diverse reasoning strategies, and refining answers through decision reflection and iterative refinement. This allows them to simulate human cognitive processes in problem-solving.

So, how do these models perform in Multilingual Machine Translation (MMT) tasks? Currently, the capabilities of o1-Like LLMs in MMT have not been systematically studied. MMT is a highly challenging task~\cite{Wang2024WhatIT,huang2024aligning,gao2024machine,wu2024perhaps,DBLP:journals/corr/abs-2410-12543}, requiring models not only to achieve semantic alignment across languages but also to ensure translation accuracy in aspects such as commonsense reasoning, historical and cultural context, and terminology.

This study focuses on the performance of o1-Like LLMs in Multilingual Machine Translation (MMT) tasks, addressing two key research questions: 1) How does the translation performance of o1-Like LLMs compare to other LLMs across different MMT tasks? 2) How do the characteristics of o1-Like LLMs impact their translation quality?

To address the first question, we evaluate multiple mainstream o1-Like LLMs. To provide a more comprehensive comparison between o1-Like LLMs and traditional LLMs in translation performance, we introduce ChatGPT~\cite{ouyang2022training}, GPT-4o~\cite{achiam2023gpt}, and DeepSeek-v3~\cite{DBLP:journals/corr/abs-2412-19437} as baseline models. We conducted a systematic analysis across five types of translation tasks, including: Flores-200~\cite{DBLP:journals/tacl/GoyalGCCWJKRGF22} for evaluating multilingual capability, Commonsense MT~\cite{he-etal-2020-box-emnlp} for testing commonsense reasoning in translation, RTT~\cite{DBLP:conf/acl/ZhangWQSWC23} for terminology constraint translation, and CultureMT~\cite{DBLP:journals/corr/abs-2305-14328} for culturally contextualized translation. Experimental results indicate that compared to traditional LLMs, o1-Like LLMs have made significant progress in multilingual translation. Compared to the GPT-4o model among LLMs, DeepSeek-R1 performed better in the contextless scenario of the Commonsense task. Additionally, in the historical and cultural task, o1-Like LLMs demonstrated stronger understanding and the ability to generate accurate cultural information. However, we also found that o1-Like LLMs centered on Chinese exhibit \textbf{Rambling} issues in cross-linguistic translation.

For the second question, we analyzed certain characteristics of o1-Like LLMs in translation tasks and examined their significant differences from traditional LLMs in multiple aspects. 
First, compared to traditional LLMs, o1-Like LLMs have higher inference costs and exhibit significantly slower inference speeds. This means that handling complex translation tasks requires more time and more computational resources. Second, translation quality shows a generally positive correlation with model size.  Finally, we found that the temperature parameter has a particularly significant impact on translation results. In complex linguistic environments, a lower temperature generally leads to more stable and accurate translations. These findings not only reveal the characteristics of o1-Like LLMs in multilingual translation tasks but also provide valuable insights for further optimizing their translation performance. The main contributions of this study are as follows:

\begin{itemize}[topsep=1pt, partopsep=-1pt, itemsep=-1pt]
    \item We evaluate multiple mainstream o1-Like LLMs across six benchmark tests, covering different dimensions of multilingual machine translation tasks, providing a comprehensive analysis of their translation capabilities.

    \item We systematically compare the translation results of o1-Like LLMs with traditional LLM-based translation systems, revealing the performance differences between the two paradigms and their potential implications.
    
    \item We identify and summarize new patterns of o1-Like LLMs in MMT tasks, exploring their advantages and limitations in depth, offering insights for future optimization.
    
\end{itemize}

\section{Related Work}

\textbf{o1-Like LLMs} Recently, o1-Like LLMs have shown exceptional performance in reasoning tasks, especially in mathematics and coding. Following OpenAI's O1 model~\cite{openai2024reasoning}, significant efforts have been made to replicate its success. \citealp{DBLP:journals/corr/abs-2410-18982} introduced journey learning, a training paradigm that enhances long-term reasoning with only 327 training samples. \citealp{DBLP:journals/corr/abs-2411-16489} demonstrated the effectiveness of data distillation from existing o1-Like models. \citealp{DBLP:journals/corr/abs-2411-14405} proposed the Marco-o1 model, combining Chain-of-Thought (CoT) fine-tuning, Monte Carlo Tree Search (MCTS), and reflection mechanisms to tackle open-ended problems. \citealp{guo2025deepseek} introduced the DeepSeek-R1 model, enhancing reasoning through multi-stage training and reinforcement learning (RL). QwQ~\cite{qwen2024qwq} model based on the Qwen architecture excel in mathematics and coding tasks but facing challenges like language mixing and circular reasoning. \citealp{DBLP:journals/corr/abs-2412-17498} proposed the DRT-o1 model, applying long CoT to MT, showing superior translation capabilities, especially with literature texts involving metaphors and similes.

\textbf{Machine Translation with Large Language Models (LLM-MT)}. Large language models, such as ChatGPT~\cite{ouyang2022training}, have shown significant effectiveness in machine translation across various language pairs~\cite{hendy2023good,jiao2023chatgpt,le2023bloom,DBLP:conf/wmt/IyerCB23,DBLP:journals/corr/abs-2311-02851,DBLP:conf/wmt/KarpinskaI23,DBLP:conf/wmt/MoslemRMKHW23,DBLP:conf/emnlp/WangLJZY0T23,DBLP:conf/wmt/IyerCB23,DBLP:conf/emnlp/FarinhasSM23}. Recent research has explored the performance of LLMs in machine translation, including control over formality in translation outputs~\cite{garcia2022using}, in-context translation abilities during pre-training ~\cite{shin2022effect}, and the impact of LLM-based machine translation on culturally sensitive texts~\cite{DBLP:journals/corr/abs-2305-14328}. Additionally, studies have examined the bilingual capabilities of LLMs to enhance translation performance~\cite{huang2024aligning}. For translation tasks requiring reasoning, multi-agent debates can effectively enhance the reasoning abilities of LLM-MT~\cite{liang2023encouraging}. These investigations further validate the research value of LLM-MT, offering diverse research directions for scholars.

\section{Evaluation Setting}
\label{setup}

\textbf{Dataset}: We conduct experiments on four MT benchmarks: Flores-200,  RTT test data, Commonsense MT, and Culture MT. 
The Flores-200 dataset~\cite{DBLP:journals/tacl/GoyalGCCWJKRGF22}\footnote{https://github.com/facebookresearch/flores} is a multilingual translation benchmark that covers 200 languages. It provides a comprehensive evaluation of translation quality across a wide range of languages. The dataset includes parallel sentences for each language pair, allowing for a thorough assessment of translation performance.
Commonsense MT~\cite{he-etal-2020-box-emnlp}\footnote{https://github.com/tjunlp-lab/commonmt} is a commonsense reasoning MT test set that evaluates the ability of translation systems to handle sentences requiring commonsense knowledge. It includes sentences with ambiguous or context-dependent meanings, making it a rigorous benchmark for assessing the ability of translation systems to generate accurate and contextually appropriate translations.
The cultural MT dataset \cite{DBLP:journals/corr/abs-2305-14328}\footnote{https://github.com/BigBinnie/Benchmarking-LLM-based-Machine-Translation-on-Cultural-Awareness} provides a culturally relevant parallel corpus with annotations of cultural-specific items. It covers 6 language pairs: En$\leftrightarrow$Es/Fr/Hi/Ta/Te/Zh, including over 7,000 cultural-specific items from 18 concept categories across more than 140 countries and regions.
RTT test data~\cite{DBLP:conf/acl/ZhangWQSWC23} is a challenging test set specifically designed for terminology MT. It increases the average constraint count per sentence from 1.1$\sim$1.7 to 6.1 and the length per target constraint from 1.1$\sim$1.2 words to 3.4 words, making it a rigorous benchmark for evaluating the handling of specialized terms in translation.

\textbf{Evaluation Metrics:} In evaluating our translation methodology, we initially employ COMET\footnote{https://huggingface.co/Unbabel/wmt22-comet-da} \cite{rei-etal-2022-comet-reference} and BLEURT~\cite{DBLP:conf/acl/SellamDP20}\footnote{https://github.com/lucadiliello/bleurt-pytorch} as automatic metrics, aligning with the established standards in LLM-based translation \cite{He2023ExploringHT,huang2024aligning}. For traditional translation evaluation, we use BLEU~\cite{DBLP:conf/wmt/Post18}\footnote{https://github.com/mjpost/sacrebleu} .

\section{Models}
We evaluate translation performance on 6 o1-Like models. OpenAI-o1~\cite{openai2024reasoning} is a close-source LLM that improves the reasoning quality by increasing the length of CoT for the first time. OpenAI-o3-mini is a small reasoning model that excels at programming, math, and science problems. DeepSeek-R1~\cite{guo2025deepseek} is a 671B model which is first fine-tuned using a large number of CoT examples, after which it was trained with Reinforcement Learning. Marco-o1-7B~\cite{DBLP:journals/corr/abs-2411-14405} is trained based on Qwen2-7B-Instruct with multiple CoT and reasoning datasets. DRT-o1-14B~\cite{DBLP:journals/corr/abs-2412-17498} is a reasoning translation model, focusing on English-to-Chinese translation task. QwQ-32B~\cite{qwen2024qwq} performs well on math and reasoning problems.

For comparison, we selected three models. ChatGPT~\cite{ouyang2022training} is a non-o1-Like model which can interact with users in a conversational manner. GPT-4o~\cite{achiam2023gpt} is a large multimodal model which demonstrates excellent ability on various QA benchmarks. DeepSeek-v3~\cite{DBLP:journals/corr/abs-2412-19437} is a Mixture-of-Experts(MoE) model which performs well on multiple benchmarks.

\section{Experimental Results}
\label{Experimental Results}
We utilized APIs to access the models from DeepSeek and OpenAI~\footnote{The DeepSeek and OpenAI models used in this study are accessed through the DeepSeek-R1, gpt-3.5-turbo, gpt-4o-2024-11-20, o1-preview-2024-09-12, and o3-mini-2025-01-31 APIs, respectively.}. For the open-source models, we employed vLLM~\footnote{https://github.com/vllm-project/vllm} for local deployment and conducted tests on 12 A100 GPUs, each with 80GB of memory. The prompt template utilized in our study is presented as follows:
\begin{mdframed}[backgroundcolor=purple!10, linecolor=white, linewidth=2pt, roundcorner=10pt]
\small
\{"role": "user", "content": "Source: $s$, translate the following sentences from $L_s$ to $L_t$"\}
\end{mdframed}
In the template, $L_s$ and $L_t$ denote the names of source language and the target language, respectively, while $s$ is the source sentence which is to be translated.


\begin{table*}[!ht]\centering
\centering
\resizebox{\textwidth}{!}{ 
\setlength{\tabcolsep}{3.3pt} 
\begin{tabular}{@{}lcccccccccccc}
\toprule
\textbf{Methods}                  & \multicolumn{3}{c}{\textbf{En$\rightarrow$De}}                           & \multicolumn{3}{c}{\textbf{En$\rightarrow$Ro}}                                         & \multicolumn{3}{c}{\textbf{Zh$\rightarrow$Ro}}                              & \multicolumn{3}{c}{\textbf{En$\rightarrow$Zh}}                                                            \\ \midrule
\multicolumn{1}{c}{}              & BLEU& COMET                          & BLEURT& BLEU& COMET                          & BLEURT& BLEU& COMET                & BLEURT& BLEU& COMET                & BLEURT\\ \midrule
\multicolumn{13}{c}{\textbf{Non-o1-Like LLMs}} \\ 
GPT-4o&        21.74& 68.35& 54.53& \multicolumn{1}{l}{24.91} & 74.11& 68.08& \multicolumn{1}{l}{20.77}          & 81.40& 71.41& \multicolumn{1}{l}{40.02}          & 81.24& 64.50\\
ChatGPT&        36.32& 83.65& 70.47& \multicolumn{1}{l}{38.38} & 88.57& 79.22& \multicolumn{1}{l}{20.76}          & 85.09& 73.60& \multicolumn{1}{l}{41.99}          & 86.00& 70.29\\
DeepSeek-v3&        21.67& 70.16& 57.76& \multicolumn{1}{l}{27.69} & 79.99& 73.63& \multicolumn{1}{l}{22.84}          & 86.79& 76.95& \multicolumn{1}{l}{41.78}          & 86.71& 71.81\\
\midrule \multicolumn{13}{c}{\textbf{Close-source o1-Like LLMs}} \\ 
OpenAI-o1&        \textbf{43.90}& 85.85& 72.96& \multicolumn{1}{l}{\textbf{41.92}} & \textbf{90.31}& 80.60& \multicolumn{1}{l}{\textbf{25.40}}          & \textbf{87.95}& \textbf{78.26}& \multicolumn{1}{l}{43.93}          & 88.51& \textbf{74.07}\\
OpenAI-o3-mini&        40.31& \textbf{86.11}& \textbf{73.35}& \multicolumn{1}{l}{41.49} & 90.29& \textbf{81.41}& \multicolumn{1}{l}{21.72}          & 86.86& 76.73& \multicolumn{1}{l}{\textbf{44.76}}          & \textbf{89.16}& 74.06\\
\midrule \multicolumn{13}{c}{\textbf{Open-source o1-Like LLMs}} \\ 
Marco-o1&        11.73& 80.87& 71.83& \multicolumn{1}{l}{15.33} & 77.14& 67.08& \multicolumn{1}{l}{5.34}          & 66.77& 69.78& \multicolumn{1}{l}{40.13}          & 74.53& 59.84
\\
DRT-o1-14B&        9.40& 65.97& 51.17& \multicolumn{1}{l}{19.91} & 80.08& 69.37& \multicolumn{1}{l}{2.42}          & 60.93& 50.25& \multicolumn{1}{l}{35.89}          & 57.05& 40.20\\
DeepSeek-R1&        39.44& 85.32& 72.55& \multicolumn{1}{l}{37.95} & 86.97& 77.62& \multicolumn{1}{l}{21.33} & 83.55& 73.40& \multicolumn{1}{l}{41.23} & 87.44& 72.43\\
QwQ& 9.98& 80.02& 66.23& 12.40& 81.50& 70.30& 11.73& 78.60& 66.56& \multicolumn{1}{l}{39.63} & 80.37&64.63\\ \midrule
\textbf{}                         & \multicolumn{3}{c}{\textbf{De$\rightarrow$En}}                           & \multicolumn{3}{c}{\textbf{Ro$\rightarrow$En}}                                         & \multicolumn{3}{c}{\textbf{Ro$\rightarrow$Zh}}                              & \multicolumn{3}{c}{\textbf{Zh$\rightarrow$En}}                                                                                                   \\ \midrule
\multicolumn{13}{c}{\textbf{Non-o1-Like LLMs}} \\ 
GPT-4o& \textbf{43.97}& \multicolumn{1}{c}{87.20} & \multicolumn{1}{c}{78.00}      & 41.60& \multicolumn{1}{c}{87.45} & \multicolumn{1}{c}{77.07} & 26.78& \multicolumn{1}{c}{64.27} & \multicolumn{1}{c}{48.17} & 20.77& \multicolumn{1}{c}{81.40} & \multicolumn{1}{c}{71.41}     \\
ChatGPT& 43.73& \multicolumn{1}{c}{87.03} & \multicolumn{1}{c}{77.55} & 41.59& \multicolumn{1}{c}{87.78}  & \multicolumn{1}{c}{77.00} & 31.42& \multicolumn{1}{c}{78.17} & \multicolumn{1}{c}{61.36} & 27.33& \multicolumn{1}{c}{85.40} & \multicolumn{1}{c}{72.82}  \\
DeepSeek-v3& 38.87& \multicolumn{1}{c}{85.44}   & \multicolumn{1}{c}{76.39}      &  37.61& \multicolumn{1}{c}{86.37} & \multicolumn{1}{c}{76.10}        
 &   29.38& \multicolumn{1}{c}{83.72} & \multicolumn{1}{c}{68.92} 
 &  27.56& \multicolumn{1}{c}{86.23} & \multicolumn{1}{c}{74.55}     \\
\midrule \multicolumn{13}{c}{\textbf{Close-source o1-Like LLMs}} \\ 
OpenAI-o1 & 43.55& 
\multicolumn{1}{c}{\textbf{87.54}} & \multicolumn{1}{c}{\textbf{78.70}}         &  \textbf{42.82}
& \multicolumn{1}{c}{\textbf{88.79}} & \multicolumn{1}{c}{\textbf{78.64}}       &  39.13& \multicolumn{1}{c}{85.59} & \multicolumn{1}{c}{71.46} &                               \textbf{28.96}& \multicolumn{1}{c}{\textbf{86.57}} & \multicolumn{1}{c}{\textbf{75.25}}     \\
OpenAI-o3-mini&        42.70& \multicolumn{1}{c}{87.34}           & \multicolumn{1}{c}{78.03}           &                      42.12& \multicolumn{1}{c}{88.35}           & \multicolumn{1}{c}{77.79}           &                               \textbf{39.42}& \multicolumn{1}{c}{\textbf{86.76}} & \multicolumn{1}{c}{\textbf{71.50}} &                               27.48& \multicolumn{1}{c}{86.28} & \multicolumn{1}{c}{74.41}     \\
\midrule \multicolumn{13}{c}{\textbf{Open-source o1-Like LLMs}} \\ 
Marco-o1&        38.45& \multicolumn{1}{c}{86.22}           & \multicolumn{1}{c}{77.30}           &                      27.15& \multicolumn{1}{c}{85.80}           & \multicolumn{1}{c}{75.37}           &                               25.68& \multicolumn{1}{c}{77.08} & \multicolumn{1}{c}{64.83} &                               18.98& \multicolumn{1}{c}{77.46} & \multicolumn{1}{c}{63.27}     \\
DRT-o1-14B&        18.40& \multicolumn{1}{c}{76.81}           & \multicolumn{1}{c}{66.66}           &                      22.95& \multicolumn{1}{c}{81.95}           & \multicolumn{1}{c}{70.96}           &                               19.43& \multicolumn{1}{c}{57.72} & \multicolumn{1}{c}{43.46} &                               3.54& \multicolumn{1}{c}{60.33} & \multicolumn{1}{c}{46.09}     \\
DeepSeek-R1&        40.91& \multicolumn{1}{c}{84.77}           & \multicolumn{1}{c}{74.71}           &                      40.60& \multicolumn{1}{c}{86.05}           & \multicolumn{1}{c}{75.00}           & {38.13}                     & \multicolumn{1}{c}{79.76} & \multicolumn{1}{c}{64.16} & 27.05& \multicolumn{1}{c}{85.56} & \multicolumn{1}{c}{73.43}     \\ 
 QwQ& 13.09& \multicolumn{1}{c}{83.07}& \multicolumn{1}{c}{72.65}& 16.55& \multicolumn{1}{c}{84.06}& \multicolumn{1}{c}{72.81}& 19.94& \multicolumn{1}{c}{75.75}& \multicolumn{1}{c}{64.75}& 15.91& \multicolumn{1}{c}{82.32}& \multicolumn{1}{c}{69.33}\\\bottomrule
\end{tabular}}
\caption{The main results from the Flores-200 dataset are presented. The bold indicates the highest values that are statistically significant, with p-values less than 0.01 in the paired t-test against all compared methods.}

\label{floresdata}
\end{table*}

\subsection{Flores-200: Evaluation of Multilingual Translation Performance}

\textbf{Main Result}: In order to evaluate multilingual translation capability, we tested the models on Flores-200 dataset. The results are prensented in Table \ref{floresdata}. In multilingual translation tasks, closed-source o1-Like LLMs demonstrated the best performance, particularly OpenAI o1, which achieved a maximum BLEU score increase of 34.5. On average, this category of models outperformed other model types by a BLEU score increase of 11.14. Among open-source models, DeepSeek-R1 performed the best, achieving an average BLEU score increase of approximately 16.92 compared to other open-source LLMs. For open-source models with relatively smaller parameter sizes, such as Marco-o1 and DRT-o1, which have 7B and 14B parameters respectively, their performance in terms of COMET and BLEURT metrics across multiple translation directions approached that of closed-source models. \textit{In the future, leveraging small-parameter open-source o1-Like LLMs for multilingual translation represents a promising research direction.} 

\textbf{Metrics Discussion}: During our experiments, we observed that the o1-Like model exhibited a much more pronounced increase in scores on COMET and BLEURT than on BLEU. In some datasets, the COMET and BLEURT scores of the o1-Like model were on par with or even exceeded those LLMs, whereas its BLEU scores were significantly lower than those LLMs. This phenomenon was especially evident in the QwQ. For example, in the En->De task of the FLORES-200 dataset, QwQ surpassed both GPT-4o and DeepSeek-v3 in COMET and BLEURT, only trailing ChatGPT by about 4. However, in BLEU, QwQ scored lower than all three non-o1-Like models, with gaps exceeding 10 in each case. This raised important questions about the unique translation characteristics of o1-Like models and how best to evaluate their performance.

As is well-known, BLEU is based on n-gram lexical matching, whereas COMET and BLEURT are designed to capture deeper semantic and contextual elements of sentences. These metrics align more closely with human judgment. For o1-Like models, the ability to engage in deeper reasoning leads to more varied expressions, sometimes using different vocabulary or sentence structures from the reference translation, while still preserving meaning. This diversity is not seen as a drawback by COMET and BLEURT, but rather to offer a more objective evaluation.

\subsection{Commonsense MT: Evaluation of Reasoning Tasks in Translation}
\begin{table*}[!ht]\centering
\centering
\resizebox{\textwidth}{!}{ 
\setlength{\tabcolsep}{3.3pt} 
\begin{tabular}{@{}lccccccccc}
\toprule
\textbf{Methods}                  & \multicolumn{3}{c}{\textbf{Lexical}}                           & \multicolumn{3}{c}{\textbf{Contextless}}                                         & \multicolumn{3}{c}{\textbf{Contextual}}                                                                                          \\ \midrule
\multicolumn{1}{c}{}              & BLEU& COMET& BELURT& BLEU& COMET& BELURT& BLEU& COMET& BELURT\\ 
\midrule\multicolumn{10}{c}{\textbf{Non-o1-Like LLMs}} \\ 
GPT-4o&        \textbf{30.57}& 80.72& 68.14& \multicolumn{1}{l}{\textbf{29.20}} & 81.34& 69.54& \multicolumn{1}{l}{\textbf{31.89}}          & 83.00& 70.34\\
ChatGPT&        24.42& 78.72& 68.61& \multicolumn{1}{l}{25.74} & 79.26& 68.83& \multicolumn{1}{l}{30.13}          & \textbf{85.09}& \textbf{73.60}\\
DeepSeek-v3&        28.53& 81.17& 69.94& \multicolumn{1}{l}{16.57} & 79.94& 68.59& \multicolumn{1}{l}{26.45}          & 82.78&           70.74\\
\midrule\multicolumn{10}{c}{\textbf{Close-source o1-Like LLMs}} \\ 
OpenAI-o1&        26.45& \textbf{82.72}& \textbf{72.03}& \multicolumn{1}{l}{  25.17} & 81.89& \textbf{71.60}& \multicolumn{1}{l}{    25.91}          & 82.88& 70.64\\
OpenAI-o3-mini&        27.10& 81.13& 70.08& \multicolumn{1}{l}{23.43} & 78.91& 68.62& \multicolumn{1}{l}{26.74}          & 82.51& 69.78          \\
\midrule\multicolumn{10}{c}{\textbf{Open-source o1-Like LLMs}} \\ 
Marco-o1&        13.44& 78.00& 65.70& \multicolumn{1}{l}{22.06} & 77.42& 63.99& \multicolumn{1}{l}{23.91} & 81.78&  69.78\\

 DRT-o1-14B& 5.84& 69.28& 50.67& 4.50& 67.98& 47.44& 2.99& 64.13&47.29\\
 DeepSeek-R1& 27.27& 81.81& 70.94& 24.95& \textbf{82.01}& 71.15& 26.76& 83.02&71.07\\
 QwQ& 1.89& 73.50& 61.41& 1.70& 62.36& 43.46& 5.18& 75.12&61.64\\\bottomrule\end{tabular}}
\caption{The main results from the Commonsense MT dataset are presented. The bold indicates the highest values that are statistically significant, with p-values less than 0.01 in the paired t-test against all compared methods.}

\label{commonsensedata}
\end{table*}

\textbf{Main Result}: To evaluate the performance of o1-Like LLMs on commonsense reasoning translation tasks, we conducted tests using the Commonsense MT dataset. The results are presented in  Table \ref{commonsensedata}. In the Lexical task, OpenAI-o1 continued to demonstrate exceptional performance, outperforming GPT-4o on both the COMET and BLEURT metrics, with improvements of 2.00 in COMET and 3.89 in BLEURT. However, in the Contextless and Contextual tasks, traditional LLMs outperformed o1-Like LLMs. \textit{Through case analysis, we found that the lack of contextual information in the source text led o1-Like LLMs to generate translations with significant hallucinations during the thinking process, whereas traditional LLMs, unaffected by such internal reasoning, produced more reliable results.} For commonsense reasoning translation tasks, it is crucial to design effective external modules to mitigate hallucinations generated during the model's reasoning process.
\subsection{Culture MT: Analyzing Culture-Specific Translation}
\begin{table*}[!ht]
	\centering
	\resizebox{\textwidth}{!}{ 
		\setlength{\tabcolsep}{3.3pt} 
		\begin{tabular}{lcccccccccccccccccc}
			\toprule
			\textbf{Methods}                  
			& \multicolumn{3}{c}{\textbf{En$\rightarrow$ Es}}                           
			& \multicolumn{3}{c}{\textbf{En$\rightarrow$ Fr}}                                         
			& \multicolumn{3}{c}{\textbf{En$\rightarrow$ Hi}} 
			& \multicolumn{3}{c}{\textbf{En$\rightarrow$ Ta}} 
			& \multicolumn{3}{c}{\textbf{En$\rightarrow$ Te}} 
			& \multicolumn{3}{c}{\textbf{En$\rightarrow$ Zh}}      
			\\ \midrule\multicolumn{19}{c}{\textbf{Non-o1-Like LLMs}} \\
			\multicolumn{1}{c}{} & BLEU & COMET & BLEURT& BLEU & COMET & BLEURT& BLEU & COMET & BLEURT& BLEU & COMET & BELURT & BLEU & COMET & BLEURT& BLEU & COMET & BLEURT\\ \midrule
			
			GPT-4o & 27.80 & 72.00 & 56.76 & 19.64 & 68.12 & 43.80 & 16.34 & 66.71 & 53.64 & 4.23 & 70.32 & 60.57 & 5.71 & 71.63 & 57.10 & 31.26 & 76.56 & 56.02 \\
			ChatGPT & 36.39 & 79.05 & 64.78 & 24.60 & 73.26 & 49.66 & 13.64 & 68.47 & 56.93 & 2.98 & 64.16 & 53.28 & 3.54 & 65.51 & 49.64 & 28.30 & 78.07 & 59.01 \\
			DeepSeek-v3 & 23.45 & 68.70 & 54.99 & 16.23 & 64.32 & 39.46 & 9.68 & 58.28 & 46.44 & 2.83 & 63.07 & 58.20 & 2.20 & 58.05 & 49.69 & 28.60 & 74.24 & 56.81 \\
            \midrule\multicolumn{19}{c}{\textbf{Close-source o1-Like LLMs}} \\
			OpenAI-o1& 41.42 & 78.30 & 62.71 & 24.64 & 72.94 & 48.34 & \textbf{20.72} & \textbf{73.67} & \textbf{62.03} & \textbf{8.48} & \textbf{79.79} & 67.12 & \textbf{8.06} & \textbf{79.82} & \textbf{61.12} & 32.10 & 82.60 & 63.82 \\
            OpenAI-o3-mini& \textbf{41.87}& \textbf{81.15}& 66.51 & \textbf{29.15}& \textbf{76.18}& 53.02 & 17.29 & 73.17 & 61.56 & 5.74 & 78.61 & \textbf{67.14} & 5.95 & 74.30 & 58.47 & \textbf{33.22}& \textbf{83.41}& 64.67 \\
            \midrule\multicolumn{19}{c}{\textbf{Open-source o1-Like LLMs}} \\
			Marco-o1 & 30.97 & 80.09 & \textbf{67.06}& 24.57 & 74.89 & \textbf{54.92}& 0.20 & 33.86 & 22.06 & 0.51 & 38.76 & 33.31 & 0.25 & 36.32 & 29.15 & 19.03& 72.98& 56.08\\
			DeepSeek-R1 & 38.09 & 79.42 & 64.63 & 26.67 & 73.77 & 50.17 & 17.57 & 69.07 & 57.28 & 5.96 & 75.29 & 60.28 & 5.06 & 65.94 & 48.97 & 30.18 & 82.57 & \textbf{64.98}\\
            QwQ & 10.68 & 74.93 & 59.23 & 4.55 & 69.91 & 45.56 & 7.14 & 64.76 & 52.75 & 0.26 & 55.44 & 47.79 & 0.35 & 53.81 & 39.87 & 13.24 & 70.78 & 51.08\\
			\midrule
			\textbf{Methods}                  
			& \multicolumn{3}{c}{\textbf{Es$\rightarrow$ En}}                           
			& \multicolumn{3}{c}{\textbf{Fr$\rightarrow$ En}}                                         
			& \multicolumn{3}{c}{\textbf{Hi$\rightarrow$ En}} 
			& \multicolumn{3}{c}{\textbf{Ta$\rightarrow$ En}} 
			& \multicolumn{3}{c}{\textbf{Te$\rightarrow$ En}} 
			& \multicolumn{3}{c}{\textbf{Zh$\rightarrow$ En}}      
			\\ \midrule\multicolumn{19}{c}{\textbf{Non-o1-Like LLMs}} \\
			\multicolumn{1}{c}{} & BLEU & COMET & BLEURT& BLEU & COMET & BLEURT& BLEU & COMET & BLEURT& BLEU & COMET & BLEURT& BLEU & COMET & BLEURT& BLEU & COMET & BLEURT\\ \midrule
			
			GPT-4o & 41.52 & 78.69 & 63.38 & 29.83 & 74.04 & 55.55 & \textbf{34.80}& \textbf{82.08}& \textbf{64.96}& 23.48 & 75.73 & 56.53 & \textbf{25.40} & \textbf{75.73} & \textbf{56.53} & \textbf{30.94} & 80.47 & 65.12 \\
			ChatGPT & 40.79 & 79.21 & 64.18 & \textbf{31.88} & 74.15 & 55.73 & 29.75 & 81.21 & 62.70 & 18.12 & 73.80 & 54.16 & 18.12 & 73.80 & 54.16 & 30.01 & 78.77 & 61.52 \\
			DeepSeek-v3 & 34.49 & 77.17 & 62.32 & 26.62 & 72.49 & 54.76 & 25.94 & 79.90 & 62.55 & 13.28 & 71.14 & 54.35 & 13.28 & 71.14 & 54.35 & 26.34 & 78.89 & 63.78 \\
            \midrule\multicolumn{19}{c}{\textbf{Close-source o1-Like LLMs}} \\
			OpenAI-o1& \textbf{46.86} & 79.64 & 63.81 & 29.77 & 73.93 & 57.14 & 30.41 & 81.85 & 64.03 &  21.45 & \textbf{79.58} & \textbf{64.00} & 19.72 & 73.45 & 53.85 & 23.05 & \textbf{82.10} & \textbf{69.97} \\
            OpenAI-o3-mini& 43.99 & \textbf{80.53}& 65.71 & 31.03 & \textbf{75.27} & 57.21 & 30.60 & 81.12 & 63.08 & \textbf{24.22}& 75.51 & 56.24 & 24.22 & 75.51 & 56.24 & 28.52 & 80.39 & 64.29 \\
            \midrule\multicolumn{19}{c}{\textbf{Open-source o1-Like LLMs}} \\
			Marco-o1 & 37.67 & 80.32 & \textbf{67.51}& 28.41 & 75.40 & \textbf{60.14} & 2.29& 45.63& 13.99& 1.14& 45.75& 15.27& 1.60 & 45.54 & 17.99 & 0.12& 70.84& 53.91\\
			DeepSeek-R1 & 39.21 & 77.53 & 62.26 & 26.49 & 73.06 & 55.29 & 30.53 & 80.09 & 62.19 & 20.61 & 76.00 & 56.70 & 22.98 & 75.58 & 56.21 & 28.34 & 80.13 & 64.81 \\
            QwQ & 10.30 & 75.33 & 60.20 & 6.04 & 69.81 & 51.52 & 11.55 & 78.18 & 59.05 & 5.34 & 71.53 & 51.65 & 3.88& 70.69 & 50.81 & 7.68 & 74.29 & 56.60\\
			\bottomrule
		\end{tabular}
	}
	\caption{The main results from the Culture MT dataset are presented. The bold indicates the highest values that are statistically significant, with p-values less than 0.01 in the paired t-test against all compared methods.}
	\label{culturedata}
\end{table*}
 \textbf{Main Result}: To evaluate the performance of models in translating culturally specific content, we utilized the Culture MT dataset for testing. The results are presented in Table \ref{culturedata}. The results revealed that in translation tasks with English as the source language, o1-Like LLMs achieved a maximum average improvement of 4.71 in BLEU, 6.88 in COMET, and 7.23 in BLEURT compared to GPT-4o. In tasks with English as the target language, the performance of the three types of models varied. The open-source model Marco-o1, despite having only 7B parameters, demonstrated exceptional performance on the BLEURT metric, achieving a maximum improvement of 1.80 compared to OpenAI-o1. Through case analysis, we observed that o1-Like LLMs naturally incorporated the localization and comprehension of proper terminology during the thinking process, resulting in more accurate translations of proper terminology and culturally authentic expressions.

\begin{table}[!ht]
\centering

\setlength{\tabcolsep}{3.3pt} 
\resizebox{\linewidth}{!}{
\begin{tabular}{@{}lcccccc@{}}
\toprule
\textbf{Methods} & \multicolumn{3}{c}{\textbf{En$\rightarrow$De}}& \multicolumn{3}{c}{\textbf{De$\rightarrow$En}}\\ \midrule
 & BLEU & COMET & BLEURT & BLEU & COMET & BLEURT \\ 
 \midrule\multicolumn{7}{c}{\textbf{Non-o1-Like LLMs}} \\
GPT-4o & 27.10 & 73.12 & 55.30 & \textbf{46.30}& \textbf{84.40} & \textbf{74.73} \\
ChatGPT & 37.58& 80.81 & 65.10 & 45.13 & 83.66 & 73.78 \\
DeepSeek-v3 & 36.66 & 82.04 & 66.38 & 43.16 & 84.03 & 74.19 \\
\midrule\multicolumn{7}{c}{\textbf{Close-source o1-Like LLMs}} \\
OpenAI-o1 & \textbf{38.64}& 81.56& 65.87& 46.20& 81.38& 70.86\\
OpenAI-03-mini & 35.79 & \textbf{82.88} & \textbf{67.33} & 42.41 & 83.95 & 74.09 \\
\midrule\multicolumn{7}{c}{\textbf{Open-source o1-Like LLMs}} \\
Marco-o1 & 16.95 & 77.44 & 66.20 & 36.48 & 82.75 & 74.42 \\
DRT-o1-14B & 9.27& 56.11& 41.44& 21.54& 72.20& 62.79\\
DeepSeek-R1 & 35.22 & 82.47 & 67.03 & 43.20 & 83.73 & 73.61 \\
QwQ & 10.04 & 77.86 & 60.46 & 13.60 & 79.37 & 67.56   \\ \bottomrule
\end{tabular}}
\caption{The main results from the RTT dataset are presented. The bold indicates the highest values that are statistically significant, with p-values less than 0.01 in the paired t-test against all compared methods.}
\label{rttdata}
\end{table}
\subsection{RTT: Proper Terminology Translation Performance}
 \textbf{Main Result}: To assess the capability of o1-Like LLMs in translating proper terminology, we employed the RTT dataset for our experiments. The results are presented in Table \ref{rttdata}. On this dataset, non-o1-Like LLMs demonstrated strong performance, with ChatGPT achieving an average improvement of approximately 7.67 in COMET and 8.49 in BLEURT across two tasks. We observed that o1-Like LLMs (DeepSeek-R1) often generated erroneous information during the reasoning process, which adversely affected the final translation outcomes. \textit{Designing external knowledge structures to enhance the proper noun translation performance of o1-Like LLMs represents a promising research direction.} 
\subsection{Inference-time Cost Analysis}
\textbf{Experiment Setting}: The reasoning process of o1-Like LLMs is notably extended, which inevitably impacts the inference efficiency of LLMs. The critical question is whether this trade-off in efficiency results in superior performance. 

To further evaluate the efficiency of o1-Like models, we conducted a comparative analysis of inference-time costs between o1-Like models and non-o1-Like models. The experiment was performed using the lexical task from CommonsenseMT dataset. We measured the average number of tokens generated by each model and their respective generation speeds to represent the inference-time cost of each model. The detailed results are presented in Table \ref{tab:cost}.
\begin{table}[!ht]
\centering

\setlength{\tabcolsep}{3.3pt} 
\resizebox{\linewidth}{!}{
\begin{tabular}{@{}lccccc@{}}
\toprule
\textbf{Methods} & \textbf{Avg Output} & \textbf{Avg Time Cost}& \textbf{BLEU} & \textbf{COMET} & \textbf{BLEURT} \\ 
 \midrule\multicolumn{6}{c}{\textbf{Non-o1-Like LLMs}} \\

GPT-4o & 42.34& -- & \textbf{30.57}& 80.72 & 68.14\\
ChatGPT & 59.82& -- & 24.42& 78.72& 68.61\\
DeepSeek-v3 & 13.89& 1.58& 28.53& 81.17&69.94\\
 \midrule\multicolumn{6}{c}{\textbf{Close-source o1-Like LLMs}} \\
OpenAI-o1 & 392.5& 8.57& 26.45& \textbf{82.72}&  \textbf{72.03}\\
OpenAI-o3-mini & 363.1& -- & 27.10& 81.13& 70.08\\
 \midrule\multicolumn{6}{c}{\textbf{Open-source o1-Like LLMs}} \\
Marco-o1 & 399.50& 26.88& 13.44& 78.00& 65.70 \\
DRT-o1-14B & 403.86& 54.59& 5.84& 69.28& 50.67\\
DeepSeek-R1 & 577.94& 42.04& 27.27& 81.81& 70.94\\
QwQ & 623.70& 13.06& 1.89& 73.50& 61.41\\

\bottomrule
\end{tabular}}
\caption{The main results of inference-time cost analysis  are presented. Average Output is measured with tokens generated per sample. Average Time Cost is measured with seconds per sample.}
\label{tab:cost}
\end{table}

\textbf{Result and Discussion}: We observe that although o1-Like LLMs demonstrate superior performance in common reasoning tasks, they incur significantly higher inference costs. o1-Like LLMs require approximately 10 times more output tokens  and 8 to 40 times more time cost compared to standard LLMs, resulting in substantially increased costs. Additionally, the "thinking" process necessitates more extensive output generation, which considerably slows down inference speed. \textit{Consequently, this trade-off makes it challenging to achieve an optimal balance between translation quality and real-time performance.}

\subsection{Instruction Following Analysis}
\textbf{Experiment Setting and Result}: A defining characteristic of o1-Like LLMs is their extended reasoning process. However, if a LLM fails to effectively adhere to instructions, the reasoning process becomes largely meaningless. 

\begin{figure}[!ht]
    \centering
    \includegraphics[width=0.95\linewidth]{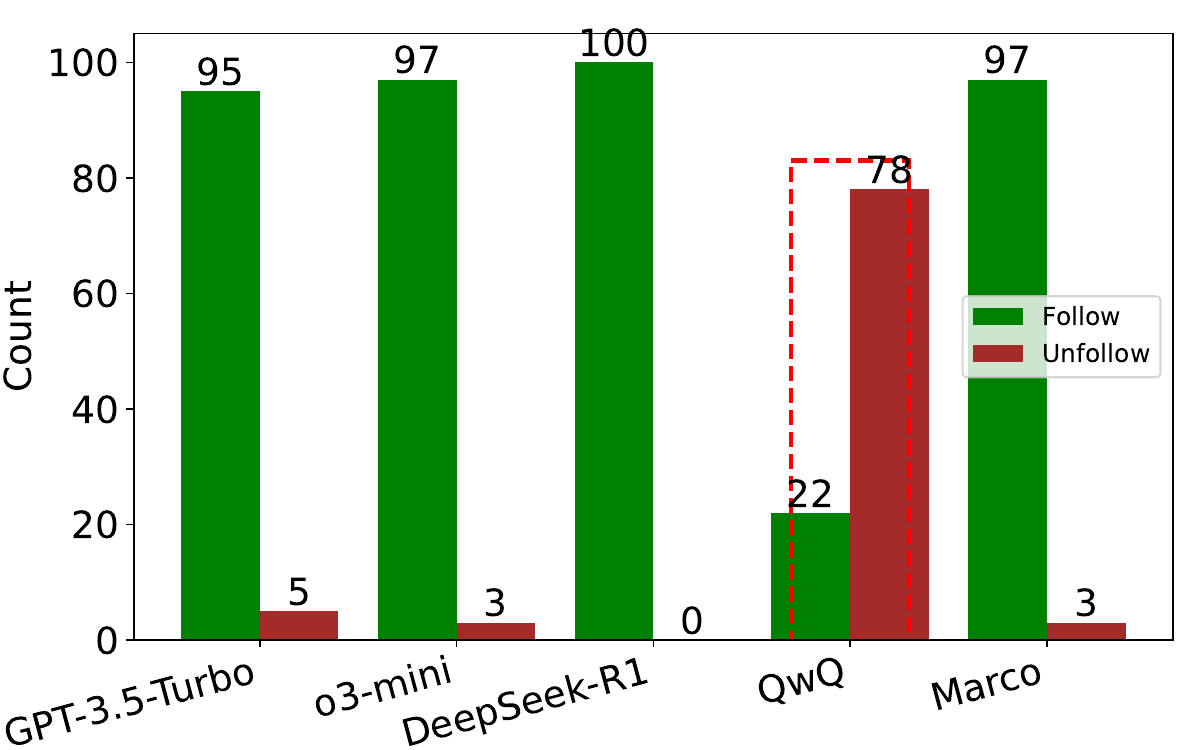}
    \caption{The main results of instruction following analysis are presented.}
    \label{fig:instruction}
\end{figure}

To assess the instruction-following capabilities of each model, we conducted the experiment using contextless tasks from the CommonsenseMT dataset. We randomly sampled 100 outputs generated by the LLMs and manually evaluated (details in Appendix \ref{huamn_evaluation}) whether they adhered to the given instructions correctly. If the model generates content that deviates from the instructions or includes redundant information in its final output, we consider that the model has not effectively adhered to the instructions. The results of this evaluation are presented in Figure \ref{fig:instruction}. 

\textbf{Discussion}: We observe that, despite being trained with complex CoT examples, o1-Like LLMs continue to encounter challenges in effectively following instructions. The probability of encountering this issue ranges approximately between 3\% and 10\%. 

Certain modules designed to maintain the instruction-following capabilities of LLMs remain essential. These modules can be utilized to further enhance the instruction-following abilities of o1-Like LLMs, representing a significant area for future research in improving the performance of o1-Like LLMs. 

Moreover, combining the results from Figure \ref{fig:instruction} and Table \ref{tab:cost}, the QwQ model exhibits poor instruction-following capabilities, often generating explanations of the source sentences rather than directly translating them. We refer to this phenomenon as “\textbf{rambling issues}”, which not only incurs higher computational overhead but also results in decreased translation quality. A case of rambling is shown in Figure \ref{ramble}, and more related cases are presented in Appendix \ref{rumblingpage}.

\begin{figure}[!h]
\centering
\begin{tcolorbox}[colback=blue!10!white, colframe=blue!50!black, width=\linewidth]
\scriptsize
\begin{itemize}
    \item \textbf{Input:} Source: 发明的是一个伟大的科学家。Please translate the following sentence from Chinese to English.
    \item \textbf{LLM Output:} So I have this task here: to translate some Chinese sentences into English. The source says it's from a great scientist. Hmm, I wonder who it is. Probably someone famous like Einstein or Newton. Anyway, I need to focus on the translation. $\cdots$ \\
Overall, I think my translations are accurate and convey the meaning of the original Chinese sentences well into English.\\
**Final Translation:**\\
\textcolor{red}{The purpose of science is to understand the workings of nature. Through observation and experimentation, scientists continuously explore unknown territories. Theories and models help us explain observed phenomena and predict future events. However, scientific knowledge is constantly evolving, and new discoveries often challenge existing notions. Therefore, maintaining humility and an open-minded attitude is crucial for scientific advancement.}
\end{itemize}
\end{tcolorbox}
\caption{An example of rambling issues is illustrated. The so-called "final translation" deviates from the exact translation, instead providing an explanation of the source text.}
\label{ramble}
\end{figure}

\textit{To some extent, striking a balance between inference speed and translation accuracy may represent a critical focus for future research.}

\subsection{Multi-scale Model Analysis}
\begin{figure*}[!ht]
    \centering
    \includegraphics[width=\linewidth]{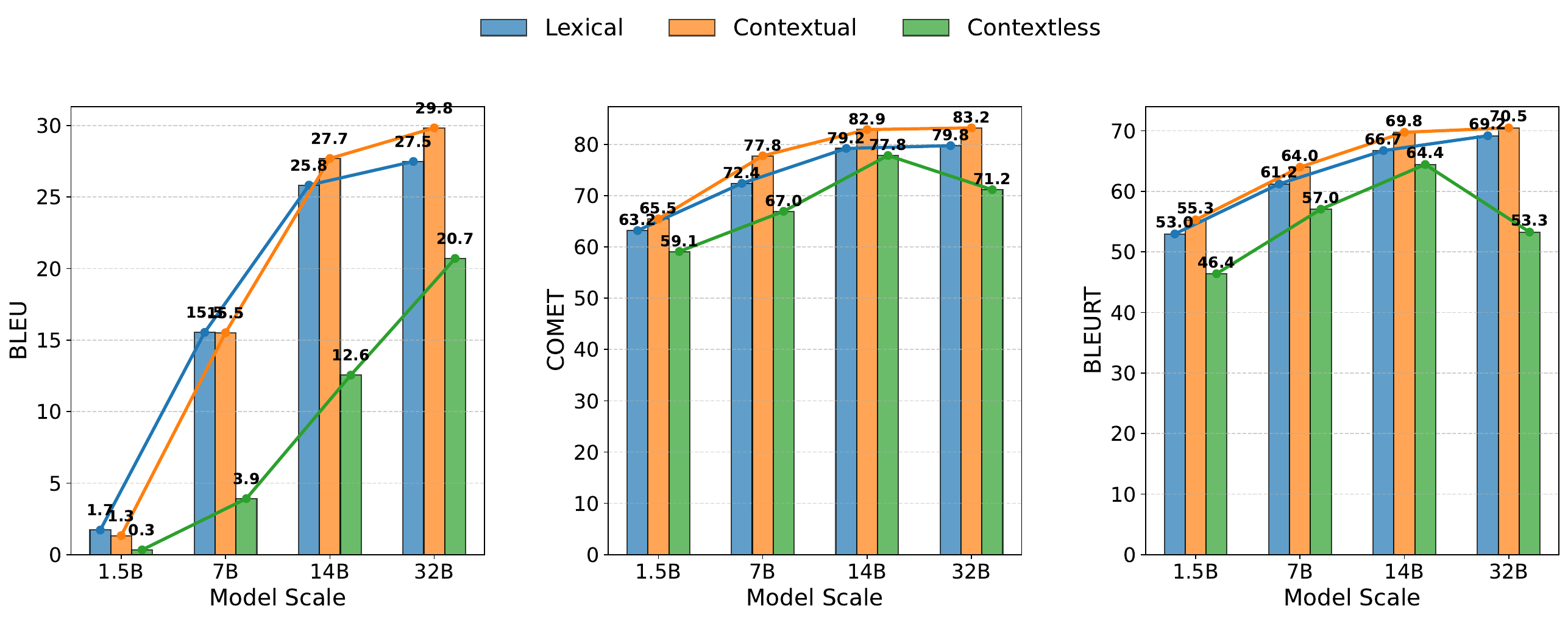}
    \caption{The main results of multi-scale model are presented.}
    \label{fig:scale}
\end{figure*}
\textbf{Experiment Setting and Result}: In non-o1-Like LLMs, it is generally observed that larger model sizes correlate with improved performance. The question arises whether o1-Like LLMs exhibit a similar characteristic. 

\begin{figure*}[!ht]
    \centering
    \includegraphics[width=1.0\linewidth]{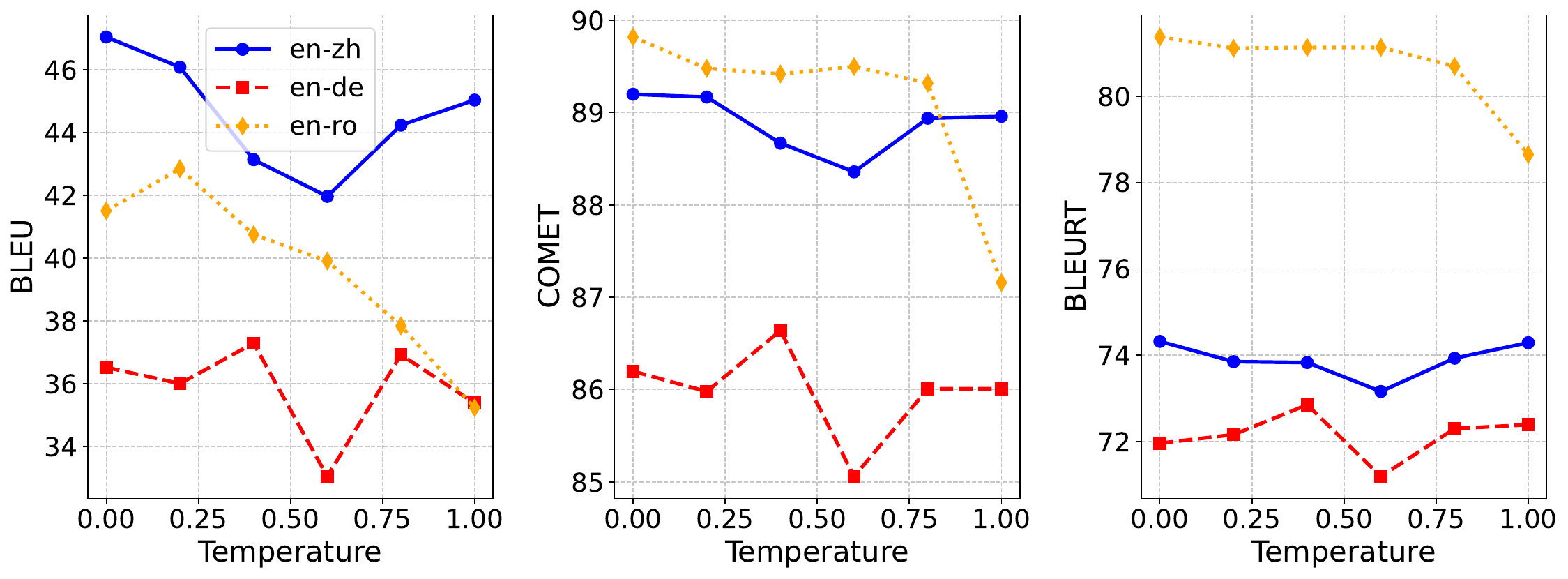}
    \caption{The main result of temperature analysis is presented.}
    \label{fig:temperature}
\end{figure*}

To investigate the relationship between the number of model parameters and translation performance, we conducted an experiment using the En$\to$De/Zh/Ro tasks of CommonsenseMT dataset. The LLM we use is DeepSeek-R1-Distill-Qwen, with a parameter of 1.5B, 7B, 14B, and 32B. The results are presented in Figure \ref{fig:scale}. 

\textbf{Discussion}: Based on the results, we observe that models with a greater number of parameters tend to exhibit better performance. However, when the number of parameters in LLMs reaches the range of 10B to 20B, further increasing on the number of parameters can only obtain marginal performance improvements. Furthermore, we observed that, in some cases, an increase in the number of parameters can lead to a decline in translation performance.
\subsection{Temperature Analysis}

\textbf{Experiment Setting and Result}: Temperature influences the probability distribution over the possible next tokens during the text generation process. A lower temperature value (e.g., close to 0) makes the model more deterministic, favoring high-probability tokens and resulting in more conservative and predictable outputs. Conversely, a higher temperature value (e.g., closer to 1 or above) increases the likelihood of selecting lower-probability tokens, thereby introducing greater variability and creativity in the generated text. 

To investigate the impact of temperature settings on the performance of LLMs, we conducted experiments using the Flores-200 dataset with the DeepSeek-R1-671B model, and the temperature sets as 0.0, 0.2, 0.4, 0.6, 0.8, and 1.0. The result is shown in Figure \ref{fig:temperature}.

\textbf{Discussion}: In general, lower temperature settings will help model produce more stable and accurate translations. Specifically, in the task of en-zh and en-de, the model's performance tends to decline and then rise as the temperature increases, with the worst performance observed around a temperature of 0.6. In contrast, the en-ro task is unique, with the model's score steadily decreasing as the temperature increases. Furthermore, the optimal temperature for the best performance varies across different tasks. This suggests that each task may have an optimal "balance point" where the model strikes the best balance between diversity and accuracy.

When considering evaluation metrics, BLEU, COMET, and BLEURT all exhibit similar trends as the temperature changes, suggesting that these metrics effectively capture the model's translation performance. However, their sensitivity to temperature variations differs. BLEU and COMET show more significant score fluctuations, while BLEURT is less affected. This difference arises from the unique characteristics of each metric, with BLEURT being more accommodating of diverse and creative translations.



\section{Conclusion}

This study evaluates o1-Like LLMs in multimodal machine translation (MMT), setting new multilingual benchmarks. DeepSeek-R1 outperforms GPT-4o in context-free tasks and excels in translating historical and cultural content, though it tends to be verbose in its Chinese outputs. Key challenges include high inference costs, slower processing, and the need for larger models for commonsense reasoning. Lower temperature settings improve stability. Future research should enhance efficiency, reduce verbosity, and refine tuning for better coherence and accuracy.

\section*{Limitations}
\label{limitations}
In this paper, we mainly evaluate the English-centric and Chinese-centric translation ability of LLMs. In the future, we aim to explore more translation directions, which will help further reveal the translation capability of o1-Like LLMs.

\bibliography{custom}

\clearpage

\appendix


\section{Human Evaluation}
\label{huamn_evaluation}
In this study, we employed a manual evaluation method to analyze the instruction-following capabilities of Large Language Models (LLMs). Since LLMs may deviate from instructions or generate redundant information during inference, relying solely on automated metrics may not accurately measure their execution of instructions. Therefore, we organized a team of three independent annotators to manually evaluate the model's outputs, ensuring the reliability and fairness of the results.

\subsection{Annotator Background and Selection}
To ensure the professionalism and consistency of the evaluation, we selected three researchers with relevant backgrounds as annotators:

\begin{itemize}
    \item \textbf{Annotator 1} -- A graduate student specializing in natural language processing, with extensive experience in large model evaluation and dataset construction, capable of accurately assessing the model’s instruction execution.
    \item \textbf{Annotator 2} -- A linguistics researcher proficient in identifying instruction-following issues within complex linguistic structures.
    \item \textbf{Annotator 3} -- A researcher focusing on LLMs, primarily responsible for ensuring that the evaluation criteria align with best practices in large model behavior assessment.
\end{itemize}

The diverse backgrounds of the annotators help analyze the model’s performance from multiple perspectives, reducing individual biases.

\subsection{Evaluation Method and Criteria}
The evaluation tasks were based on contextless tasks from the \textbf{CommonsenseMT} dataset. A random sample of \textbf{100 outputs} generated by LLMs was selected, and annotators were asked to independently assess their adherence to the given instructions. The evaluation was based on three key criteria:

\begin{enumerate}
    \item \textbf{Instruction Adherence} -- Determining whether the model’s output strictly follows the given instruction, avoiding deviations from the task requirements.
    \item \textbf{Output Consistency} -- Checking if the model's response is semantically coherent and maintains relevance to the input instruction.
    \item \textbf{Redundancy} -- Assessing whether the output contains unnecessary or extraneous information, particularly content unrelated to the instruction.
\end{enumerate}

Each annotator provided a \textbf{binary score} (1 for adherence, 0 for non-adherence) for each sample based on the above criteria. In cases where there was disagreement among the three annotators, the final score was determined by majority vote.

\subsection{Safety and Consistency Assurance}
To ensure the \textbf{safety} and \textbf{alignment} of the evaluation process, multiple measures were implemented:

\begin{itemize}
    \item \textbf{Data Anonymization} -- All evaluation data was anonymized to prevent the inclusion of any personally identifiable or sensitive information.
    \item \textbf{Detailed Annotation Guidelines} -- Clear evaluation criteria were established to minimize subjective interpretation differences and maintain consistency across annotators.
\end{itemize}

These measures help enhance the objectivity, reliability, and security of the manual evaluation process.

\section{Case Study}
\subsection{"Rambling" Case}
\label{rumblingpage}
More rambling cases are prensented in Figure \ref{rambleapp1} and \ref{rambleapp2}.
\begin{figure*}[!ht]
\begin{tcolorbox}[colback=blue!10!white, colframe=blue!50!black, width=\textwidth]
\begin{itemize}
    \item \textbf{Input:} Source: 卖多余作业本。Please translate the following sentence from Chinese to English.
    \item \textbf{LLM Ouput:} So I have this task here: I need to translate some sentences from Chinese to English. The source is about selling extra exercise books. Okay, let's see what I've got.First, I need to understand the context. It's about selling extra exercise books, which probably means someone has some unused exercise books and wants to sell them. Maybe a student who has more books than they need or a teacher clearing out supplies.I should probably start by identifying the key phrases and words in the Chinese sentences. Then, I can think about how to convey the same meaning in English.Let me assume there are a few sentences to translate. Since they're not provided, I'll create some example sentences that fit the context and translate those.Example sentence 1: "我有一些多余的作业本要卖。"Translation: "I have some extra exercise books to sell."Explanation: Here, "我有" means "I have," "一些" means "some," "多余 的" means "extra" or "surplus," "作业本" are exercise books, and "要卖" means "to sell."Example sentence 2: "这些作业本是新的，从来没有用过。"Translation: "These exercise books are new and have never been used."Explanation: "这些" means "these," "是" means "are," "新的" means "new," "从来没有" means "have never," "用过" means "been used."Example sentence 3: " 如果你感兴趣，可以联系我。"Translation: "If you're interested, you can contact me."Explanation: "如果你" means "if you," "感兴趣" means "are interested," "可以" means "can," "联系" means "contact," "我" means "me."Example sentence 4: "价格可以商量。"Translation: "The price is negotiable."Explanation: "价格" means "price," "可以" means "can be," "商量" means "negotiated."Putting it all together, the English version would be:"I have some extra exercise books to sell. These exercise books are new and have never been used. If you're interested, you can contact me. The price is negotiable."This should give a clear idea of what's being offered and how to get in touch.I think this approach covers the basics. If there are specific sentences provided, I can adjust the translations accordingly. But since they weren't included, these examples should serve as a good guide.\\
     **Final Answer**: \\
    \textcolor{red}{I have some extra exercise books to sell. These exercise books are new and have never been used. If you're interested, you can contact me. The price is negotiable.}
\end{itemize}

\end{tcolorbox}
\caption{An example of \textbf{rambling issues} is illustrated. The so-called "final translation" deviates from the exact translation, instead providing an explanation of the source text.}
\label{rambleapp1}
\end{figure*}

\begin{figure*}[!ht]
\begin{tcolorbox}[colback=blue!10!white, colframe=blue!50!black, width=\textwidth]
\begin{itemize}
    \item \textbf{Input:} Source: 卖多余作业本。Please translate the following sentence from Chinese to English.
    \item \textbf{LLM Ouput:} Okay, I have this task to translate some sentences from Chinese to English. The source says it's from a simple course, so probably basic-level sentences. I need to make sure I understand the meaning correctly and translate them accurately. Let's see.First, I need to know what the sentences are. Since they're not provided here, I'll assume some common simple Chinese sentences for practice. For example:1. 我的名字是李华。2. 我喜欢看电影。3. 今天天气很好。4. 我每天早上喝咖啡。5. 我不喜欢吃辣的食物。Alright, let's translate them one by one.Starting with the first one: "我的名字是李华。"- "我的" means "my."- "名字" means "name."- "是" means "is."- "李华" is a name, so it should be transliterated as "Li Hua."So, putting it together: "My name is Li Hua."Next, "我喜欢看电影。"- "我" means "I."- "喜欢" means "like."- "看" means "watch" or "see."- "电影" means "movie."So, "I like to watch movies." or "I like watching movies." Both are acceptable, but maybe "watching" sounds more natural.Third sentence: "今天天气很好。"- "今天" means "today."- "天气" means "weather."- "很好" means "very good."So, "Today's weather is very good." or more naturally, "The weather today is very good." But in English, we often say "Today's weather is great." So maybe adjust it to sound more natural.Fourth sentence: "我每天早上喝咖啡。"- "我" means "I."- "每天" means "every day."- "早上" means "morning."- "喝" means "drink."- "咖啡" means "coffee."So, "I drink coffee every morning." That seems straightforward.Last sentence: "我不喜欢吃辣的食物。"- "我" means "I."- "不" means "not."- "喜欢" means "like."- "吃" means "eat."- "辣的" means "spicy."- "食物" means "food."So, "I don't like to eat spicy food." or "I don't like eating spicy food." Again, both are acceptable, but maybe "eating" sounds more natural.Alright, so summarizing:1. My name is Li Hua.2. I like watching movies.3. The weather today is very good.4. I drink coffee every morning.5. I don't like eating spicy food.I think these translations are accurate and sound natural in English. It's important to not only translate word-for-word but also consider how native speakers would express those ideas.\\
    \#\# Final Translation\\
    \textcolor{red}{1. My name is Li Hua.2. I like watching movies.3. The weather today is very good.4. I drink coffee every morning.5. I don't like eating spicy food.}
\end{itemize}

\end{tcolorbox}
\caption{An example of rambling is illustrated. The so-called "final translation" deviates from the exact translation, instead providing an explanation of the source text.}
\label{rambleapp2}
\end{figure*}

\end{CJK}
\end{document}